\documentclass[letterpaper, 10 pt, conference]{ieeeconf}  % Comment this line out if you need a4paper

\IEEEoverridecommandlockouts                              % This command is only needed if 
\usepackage{graphics} % for pdf, bitmapped graphics files
\usepackage[pdftex]{graphicx}
\usepackage{amsmath} % assumes amsmath package installed
\usepackage{cite}
\usepackage{subfig}
\usepackage{float}
\usepackage{multirow}
\usepackage{tabularx}
\usepackage{mathtools}

\title{\LARGE \bf
LocNet: Global Localization in 3D Point Clouds for Mobile Vehicles
}

\author{Huan Yin,
	Li Tang,
	Xiaqing Ding,
	Yue~Wang~\IEEEmembership{Member,~IEEE}, 
	and
	Rong~Xiong~\IEEEmembership{Member,~IEEE}% <-this % stops a space
	\thanks{All authors are with the State Key Laboratory of Industrial Control and Technology, Zhejiang University, Hangzhou, P.R. China. Yue Wang is with iPlus Robotics Hangzhou, P.R. China. Yue Wang is the corresponding author {\tt\small wangyue@iipc.zju.edu.cn}.}
	\thanks{This work was supported in part by the National Nature Science Foundation of China under Grant U1609210, in part by the Science Fund for Creative Research Groups of NSFC under Grant 61621002, and in part by the Man-Machine Cooperative Mobile Dual-Arm Robot for Dexterous Operation in Intelligent robot special of national key research and development program under Grant 2017YFB1300400. }% <-this % stops a space
}

\begin{document}

\maketitle
\thispagestyle{empty}
\pagestyle{empty}

%%%%%%%%%%%%%%%%%%%%%%%%%%%%%%%%%%%%%%%%%%%%%%%%%%%%%%%%%%%%%%%%%%%%%%%%%%%%%%%%
\begin{abstract}

Global localization in 3D point clouds is a challenging problem of estimating the pose of vehicles without any prior knowledge. In this paper, a solution to this problem is presented by achieving place recognition and metric pose estimation in the global prior map. Specifically, we present a semi-handcrafted representation learning method for LiDAR point clouds using siamese LocNets, which states the place recognition problem to a similarity modeling problem. With the final learned representations by LocNet, a global localization framework with range-only observations is proposed. To demonstrate the performance and effectiveness of our global localization system, KITTI dataset is employed for comparison with other algorithms, and also on our long-time multi-session datasets for evaluation. The result shows that our system can achieve high accuracy.

\end{abstract}

%%%%%%%%%%%%%%%%%%%%%%%%%%%%%%%%%%%%%%%%%%%%%%%%%%%%%%%%%%%%%%%%%%%%%%%%%%%%%%%%
\section{Introduction}

Localization aims at estimating the pose of mobile vehicles, thus it is a primary requirement for  autonomous navigation. When a global map is given and the prior pose is centering on the correct position, pose tracking is enough to handle the localization of the robot or vehicle \cite{thrun2005probabilistic}. In case the vehicle misses its pose in GPS-denied urban environments, re-localization module tries to re-localize the vehicle. While for mapping, loop closing is a crucial component for global consistent map building, as it can localize the vehicle in the previously mapped area. Note that for both loop closing and re-localization, their underlying problems are the same, which requires the vehicle to localize itself in a given map without any prior knowledge as soon as possible. To unify the naming, we call this common problem as global localization in the remainder of this paper.

Global localization is relatively mature when the robot is equipped with a 2D laser range finder \cite{hess2016real}. For vision sensors, the topological loop closure is investigated in recent years \cite{cummins2008fab}, and the pose is mainly derived by classical feature keypoints matching. The success of deep learning raised the learning based methods as new potential techniques for solving visual localization problems \cite{chen2014convolutional}. However, for 3D point cloud sensor, e.g. 3D LiDAR, which is popular in the area of autonomous driving, there are less works focusing on the global localization problem \cite{dube2017segmatch}, which mainly focused on designing representations for matching between sensor readings and the map. Learning the representation of 3D point cloud is explored in objects detection \cite{Li-RSS-16}, but inapplicable in localization problem. In summary, the crucial challenge we consider is the lack of compact and efficient representation for global localization.

\begin{figure*}[t]
	\centering
	\includegraphics[width=14cm]{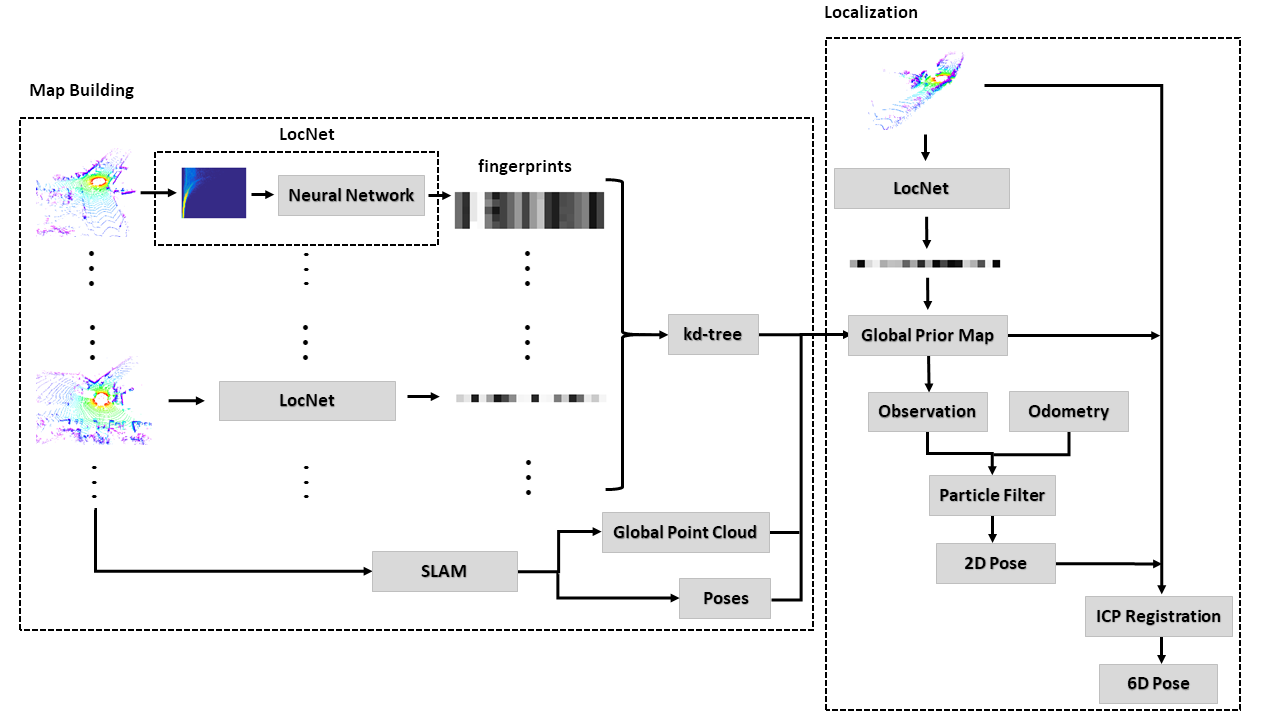}
	\caption{ The framework of our global localization system. }
	\label{fig:frame}	
\end{figure*}

In this paper, we propose a semi-handcrafted deep neural network, LocNet, to learn the representation of 3D LiDAR sensor readings, upon which, a Monte-Carlo localization framework is designed for global metric localization. The frame of our localization system is illustrated in Fig.~\ref{fig:frame}. The sensor readings is first transformed as a handcrafted rotational invariant representation, which is then passed through a network to generate a low-dimensional fingerprint. Importantly, when the representations of two readings are close, the two readings are collected in the same place with high probability. With this property, the sensor readings are organized as a global prior map, on which a particle based localizer is presented with range-only observations to achieve the fast convergence to correct the location and orientation.

The contributions of this paper are presented as follows:

\begin{itemize}
	\item A handcrafted representation for structured point cloud is proposed, which achieves rotational invariance, guaranteeing this property in the learned representation.
	\item Siamese architecture is introduced in LocNet to model the similarity between two LiDAR readings. As the learning metric is built in the Euclidean space, the learned representation can be measured in the Euclidean space by simple computation.
	\item A global localization framework with range-only observations is proposed, which is built upon the prior map made up of poses and learned representations.
\end{itemize}

The rest of this paper is organized as follows: Section
\uppercase\expandafter{\romannumeral2} describes related work of global localization. In Section 		
\uppercase\expandafter{\romannumeral3}, we introduce the details of our system. We evaluate our system on KITTI odometry benchmark and our own dataset in Section \uppercase\expandafter{\romannumeral4}. Section
\uppercase\expandafter{\romannumeral5} concludes with a brief discussion on our methods.

%%%%%%%%%%%%%%%%%%%%%%%%%%%%%%%%%%%%%%%%%%%%%%%%%%%%%%%%%%%%%%%%%%%%%%%%%%%%%%%%
\section{Related Work}

In general, the global localization consists of two stages, place recognition, which finds the frame in the map that is topologically close to the current frame, and metric pose estimation, which yields the relative pose from the map frame to the current frame, thus localizing the vehicle finally. We review the related works in two steps, place recognition and metric pose estimation.

As images provide rich information of the surrounding environments, the place recognition phase in the global localization is quite mature in vision community. Most image based localization methods employed bag of words for description of a frame \cite{cummins2008fab}. Based on this scheme, sequences of images were considered for matching instead of single frame to improve the precision \cite{milford2012seqslam}. To enhance the place recognition across illumination variance, illumination invariant description was studied for better matching performance \cite{naseer2015robust}. However, given the matched frame in the map to the current frame, the pose estimation is still challenging, since the feature points change a lot across illumination, and the image level descriptor cannot reflect the information of the metric pose.

Compared with vision based methods, the place recognition in point clouds does not suffer from various illumination. A direct method of matching current LiDAR with the given map is registration. Go-ICP, a global registration method without initialization was proposed in \cite{yang2016go}. But it is relatively computational complex. Thus we still refer to the two-step method, place recognition then metric pose estimation. Some works achieved place recognition in the semantic level. SegMatch, presented by Dube et al. \cite{dube2017segmatch}, tried to match to the map using features like buildings, trees or vehicles. With these matched semantic features, a metric pose estimation was then possible. As semantic level feature is usually environment dependent, point features in point cloud are also investigated. Spin image \cite{johnson1999using} was a keypoint based method to represent surface of 3D scene. ESF \cite{wohlkinger2011ensemble} used distance and angle properties to generate keypoints without computing normal vectors. Bosse and Zlot \cite{bosse2013place} proposed 3D Gestalt descriptors in 3D point clouds for matching. These works mainly focused on the local features in the frames for matching.

\begin{figure*}[t]
	\centering
	\includegraphics[width=14cm]{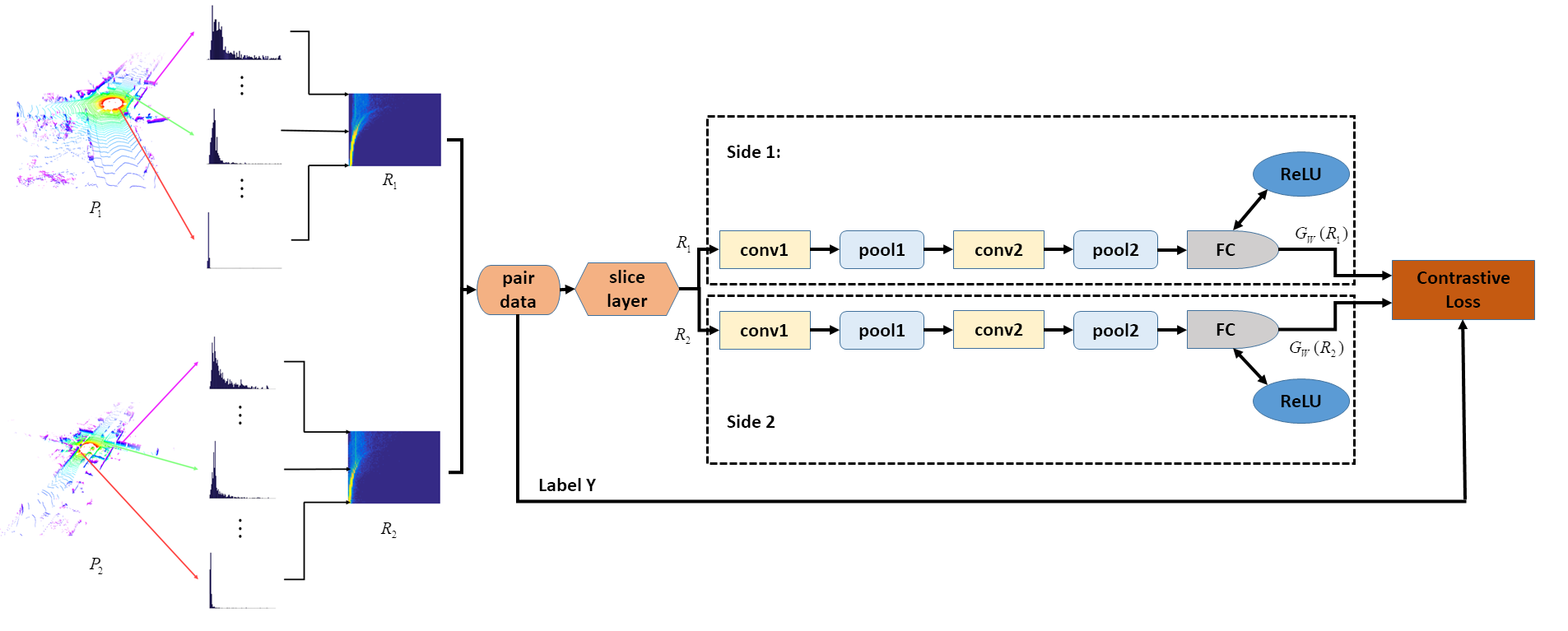}
	\caption{The framework of our siamese network for training, including the generation process of handcrafted representations.}
	\label{fig:network}	
\end{figure*}

For frame level descriptors, there were some works utilizing the handcrafted representation. Magnusson et al. \cite{magnusson2009appearance} proposed an appearance based loop detection method using NDT surface representation. R{\"o}hling et al. \cite{rohling2015fast} proposed a 1-D histogram to describe the range distribution of the entire point cloud. For learning based method, Granstr{\"o}m and Sch{\"o}n \cite{granstrom2011learning} used features that capture important geometric and statistical properties of 2D point clouds. The features are used as input to the machine learning algorithm - Adaboosts to build a classifier for matching. Based on the deep learning, our method set to develop the learning based representations for 3D point clouds.

%%%%%%%%%%%%%%%%%%%%%%%%%%%%%%%%%%%%%%%%%%%%%%%%%%%%%%%%%%%%%%%%%%%%%%%%%%%%%%%%\

\section{Methods}

The proposed global localization system includes two components as shown in Fig. 1, map building and localization. In the map building component, frames collected in the mapping session are transformed through LocNet to generate the corresponding fingerprints, forming a kd-tree based vocabulary for online matching as a global prior map. The localization component is utilized in the online session, which transforms the current frame to the fingerprint using LocNet for searching similar frames in the global prior map. One can see that the crucial module in both components is the LocNet which is shown in Fig. 2. The LocNet is duplicated to form the siamese network for training. When being deployed in the localization system, LocNet, i.e. one branch of the siamese network, is extracted for representation generation. In this section, the LocNet is first introduced, followed by the global localization system design.

\subsection{Rotational Invariant Representation}

Each point in a frame $P$ of LiDAR data is described using $(x,y,z)$ in Cartesian coordinates, which can be transformed to $(r, \theta, \phi)$ in spherical coordinates. Considering the general
3D LiDAR sensor, the elevation angle $\theta$ is actually discrete, thus the frame $P$ can be divided into multiple rings of measurements, which is denoted as $ S^{i}_{N} \in P $, where $i$ is the index of rings from 1 to $N$. In each ring, the points are sorted by azimuth angle $\phi$, of which each point is denoted by $p_{k}$, where $k$ is the index determined by $\phi$. As a result, in each LiDAR frame, each point is indexed by ring index $ S^{i}_{N}$
and $\phi$ index $k$.

Given a ring $ S^{i}_{N}$ , the 2D distance between two consecutive
points $p_{k}$ and $p_{k-1}$ is calculated as
\begin{equation}
	 d(p_{k}, p_{k-1}) =\sqrt{(x_{k} - x_{k-1})^{2} + (y_{k} - y_{k-1})^{2}} 
\end{equation}
which lies in a pre-specified range interval $d \in \left[ d_{min}, d_{max}\right] $.

For a whole ring, we calculate all the distances between each pair of consecutive points. Then, we set a constant bucket count $b$ and a range value $I=\left[ d_{min}, d_{max}\right] $, and divide $I$ into subintervals of size:
\begin{equation}
	 \Delta I_{b} = \dfrac{1}{b} \left( d_{max} - d_{min}\right)  
\end{equation}
and each bucket corresponds to one of the disjunct intervals, show as follows:
\begin{equation}
	 I_{b}^{m} = \left[ d_{min}+n\cdot\Delta I_{b}, d_{min}+\left( n+1\right)\cdot\Delta I_{b}\right]  
\end{equation}
where $n$ is the bucket index and all $d$ in the ring $ S^{i}_{N} $ can find which bucket it belongs to. So the histogram for a ring $ S^{i}_{N} $ can be written as
\begin{equation}
	 H_{b}^{i} = \left( h_{b}^{0},\cdots,h_{b}^{b-1}\right) 
\end{equation}
with
\begin{equation}
	 h_{b}^{m} = \dfrac{1}{\left| S^{i}_{N}\right| }\left| \left\lbrace p_{k}\in S^{i}_{N} : d\left( p_{k}, p_{k-1}\right) \in I_{b}^{m}\right\rbrace \right|  
\end{equation}
\par Finally, we stack $ N $ histograms $ H_{b}^{i} $ together in the order of rings from top to bottom. Then a $ N \times b $ one-channel image-like representation $ R=( H_{b}^{0}, \cdots, H_{b}^{N-1} )^{T}$ is produced from the  point cloud $P$. Using this representation, if the vehicle rotates at the same place, the representation keeps constant, thus rotational invariant.

One disturbance in global localization is the moving objects, as they may cause unpredictable changes of range distributions. By utilizing the ring information, this disturbance can be tolerated to some extent, since the moving objects usually occur near the ground, the rings corresponding to higher elevation angles are decoupled from these dynamics.

\subsection{Learned Representation}

With the handcrafted representation $R$, we transform the global localization problem to an identity verification problem. Siamese network is able to solve this problem and reduce the dimension of representations, as shown in Fig.~\ref{fig:network}.

Assume that the final output of any side (Side 1 or Side 2 in Fig.~\ref{fig:network}) of the siamese neural network is a $d$ dimensional feature vector $G_{W}(R) = \left\lbrace g_{1},\cdots,g_{d} \right\rbrace $. The parameterized distance function to be learned from the neural network between image-like representations $R_{1}$ and $R_{2}$ is $ D_{W}(R_{1},R_{2}) $, which represents the Euclidean distance between the outputs of $G_{W}$, show as follows:
\begin{equation}
D_{W}(R_{1}, R_{2}) = \|  G_{W}(R_{1}) - G_{W}(R_{2}) \|_{2} 
\end{equation}
\par In the siamese convolution neural network, the most critical  part is the contrastive loss function, proposed by Lecun et al. \cite{hadsell2006dimensionality}, show as follows:
\begin{equation}
L(Y,R_{1}, R_{2})\!=\!Y\dfrac{1}{2}(D_{W})^{2}\!+\!(1\!-\!Y)\dfrac{1}{2}{max(0, m\!-\!D_{W})}^{2} 
\end{equation}
\par As one can see, the loss function needs a pair of samples to compute the final loss. Let label $Y$ be assigned to a pair of representations $R_{1}$ and $R_{2}$: $Y=1$ if $R_{1}$ and $R_{2}$ are similar, representing the two places are close, so the two frames are matched; and $Y=0$ if $R_{1}$ and $R_{2}$ are dissimilar, representing the two places are not the same which is the most common situation in localization and mapping system. Actually, the purpose of contrastive loss is try to decrease the value of $ D_{W} $ for similar pairs and increase it for dissimilar pairs in the training step.

After the network is trained, in the testing step, we assume all places are the same to the query frame. So in order to achieve the similarity between a pair of input representations $R_{1}$ and $R_{2}$, we manually set the label $Y =1$, then the contrastive loss is as following:
\begin{equation}
	 L(R_{1},R_{2}) = \dfrac{1}{2}(D_{W})^{2}
\end{equation}
\par As one can see, if the matching of two places is real, the calculated contrastive loss should be a very low value. If not real, the loss should be a higher value, and the second part of original contrastive loss should be low to minimize $L(Y,R_{1},R_{2})$. And it is easy to judge the place recognition using a binary classifier $C_{\tau}$ with threshold $\tau$ :
\begin{equation}
	C_{\tau}(R_{1}, R_{2})=
	\begin{cases}
	\text{true ,}& \text{$D_{W} \leq \tau $}\\
	\text{false ,}& \text{$D_{W} > \tau $}
	\end{cases}
\end{equation}
where parameter $\tau$ decides the place recognition result.

In summary, the advantage of using neural network is obvious. The low-dimensional representation, the fingerprint $G_{W}(R)$, is learned in the Euclidean space, so it is convenient to build the prior map for global localization.

\subsection{Global Localization Implementation}

As previously described, global localization of mobile vehicles is based on prior maps and pose estimation. Accordingly, it is necessary to build a reliable prior global map first, especially for long-time running vehicles.

In this paper, we utilize 3D LiDAR based SLAM technology to produce the map, which provides a metric pose graph $\left\lbrace X\right\rbrace $ with each node indicating a LiDAR frame, whose corresponding representations $\left\lbrace G_{W}(R)\right\rbrace $ is also included by passing the frame through LocNet, as well as the entire corrected global point cloud $P$. So, our prior global map $ M$ is made up as follows:
\begin{equation}
	 M = \left\lbrace  P, \left\lbrace G_{W}(R)\right\rbrace, \left\lbrace X\right\rbrace\right\rbrace  
\end{equation}
\par In order to improve the searching efficiency for real-time localization, a kd-tree $ K $ is built based on set $\left\lbrace G_{W}(R)\right\rbrace $ in Euclidean space. Thus the prior global map is as follows:
\begin{equation}
	 M = \left\lbrace  P, K, \left\lbrace X\right\rbrace\right\rbrace  
\end{equation}
\par After the map is built, when a new point cloud $P_{t}$ comes at time $t$, it is firstly transformed to handcrafted representation $R_{t}$ and then to the dimension reduced feature $ G_{W}(R_{t}) $ by LocNet. Assuming the closest matching result of it in $ K $ is $R_{k} $, and the pose in memory that $R_{k}$ attached with is $ X_{k} $. Thus, the observation $Z$ of the matching can be regarded as a range only observation:
\begin{equation}
	Z =  X_{k} 
\end{equation}

We utilize Monte-Carlo Localization to estimate the 2D pose $X_{t}$ with the continuous observations. And the weight is computed based on the the distance between the particles and the observed pose according to Gaussian distribution. Particles with less weights are filtered in the re-sampling step, and those with high probability are kept. After some steps, MCL is converged to a correct orientation with range-only observations and continuous odometry.    

Based on the produced $X_{t}$, we set it as the initial value of ICP algorithm \cite{Pomerleau12comp}, which help the vehicle achieve a more precise 6D pose by registering current point cloud $ P_{t} $ with the entire point cloud $P$. In overall, the whole localization process is from coarse to fine.

%%%%%%%%%%%%%%%%%%%%%%%%%%%%%%%%%%%%%%%%%%%%%%%%%%%%%%%%%%%%%%%%%%%%%%%%%%%%%%%%
\section{Experiments}

The proposed global localization system is evaluated in two aspects: the performance of place recognition, the feasibility and convergence of localization. 

\begin{figure}[t]
	\centering
	\subfloat[running routes]{\includegraphics[width=3.0cm]{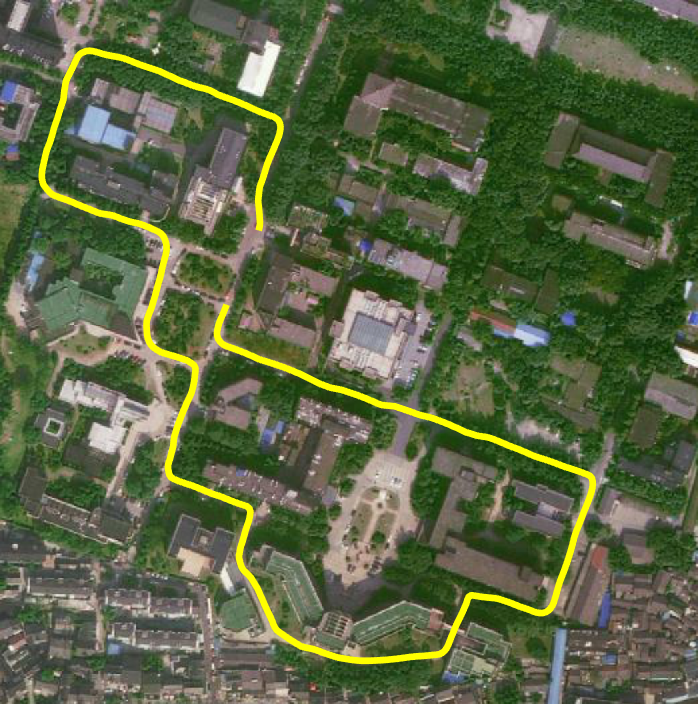}%
		\label{yq21}}
	\hfil
	\subfloat[global point cloud $ \boldsymbol{P} $]{\includegraphics[width=4.0cm]{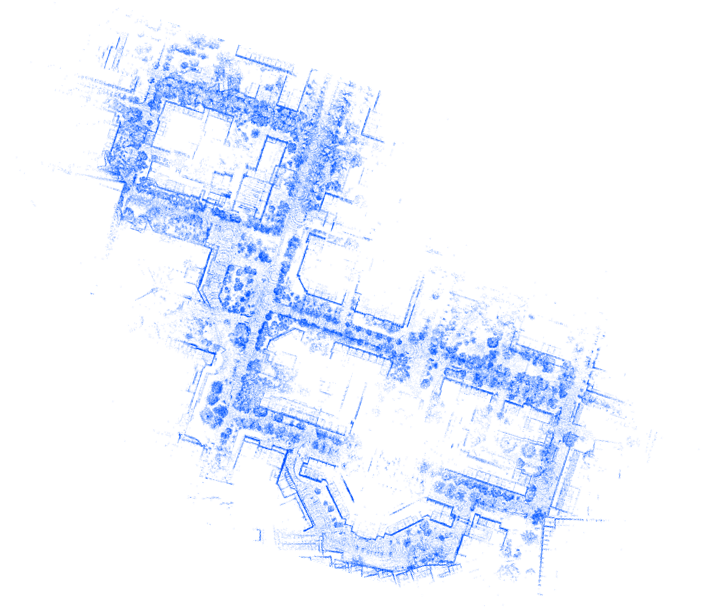}%
		\label{wholeMap}}
	\caption{(a) 6 same route lines over 3 days at the south of Yuquan campus in Zhejiang University (b) produced global point cloud $ P $ after the first running in Day 1. }
	\label{fig:YQ21dataset}
\end{figure}

\subsection{Datasets and Training}

In the experiments, two datasets are employed. First, KITTI dataset \cite{geiger2012we}, has the odometry datasets with both Velodyne HDL-64 LiDAR sensor readings and the ground truth for evaluation. We pick the odometry benchmark 00, 02, 05, 06 and 08 with loop closures for test. Second, we collect our own 21-session dataset with a length of 1.1km in each session across 3 days \cite{Tang2018}, called YQ21, from the university campus using our mobile robot. The ground truth of our datasets is built using DGPS aided SLAM and localization with handcrafted initial pose. Our vehicle equips a Velodyne VLP-16 LiDAR sensor, which also tests the feasibility to different 3D LiDAR sensors. We select 6 sessions over 3 days for evaluation, as shown in Fig.~\ref{fig:YQ21dataset}.

On different datasets, different strategies are applied to train LocNet. In KITTI dataset, we generate 11000 positive samples and 28220 negative samples from other sequences using dense poses, and set margin value $m=12$ in the contrastive loss. While in YQ21 dataset, the first session in Day 1 is used to build the target global map $ \boldsymbol{M} $. The 3 sessions in Day 2 are used to train LocNet and the last two sessions in Day 3: Day 3-1 and Day 3-2 are used to test the LocNet. The former is collected in the morning and the latter is in the afternoon, verifying that no illumination variance can affect the localization performance when using point clouds. And we finally generate 21300 positive samples and 40700 negative ones for training and set $m=8$. We implement our siamese network using caffe\footnote{http://caffe.berkeleyvision.org/}. 

\subsection{Place Recognition Performance}

We compare LocNet with other three algorithms, mainly the 3D point cloud keypoints based method, and the frame descriptor based method: Spin Image \cite{johnson1999using}, ESF \cite{wohlkinger2011ensemble} and Fast Histogram \cite{rohling2015fast}. Spin Image and ESF are based on local descriptors, and we transform the entire point cloud as a global descriptor, and the LiDAR sensor is the center of the descriptor. We use the C++ implementation of them in the PCL\footnote{http://pointclouds.org/}. As for Fast Histogram and LocNet, the buckets of histograms are set with the same value $b=80$.

In some papers, the test set is manually selected \cite{magnusson2009appearance} \cite{granstrom2011learning}, while we select to compute the performance by exhaustively matching any pairs of frame in the dataset. Based on the vectorized outputs of algorithms, we generate similarity matrix using kd-tree, then evaluate performance by comparing to the ground truth based similarity matrix. It's a huge calculation, almost half of the 4541$\times$4541 comparison times of sequence 00 for instance.

\begin{table}[t]
	\begin{center}
		\caption{ F$_{1}$max score on sequence 00 with different $p$}
		\begin{tabular}{ccccc}
			\hline
			Methods & $p=2m$& $p=3m$& $p=5m$& $p=10m$ \\
			\hline
			Spin Image& 0.600& 0.506& 0.388& 0.271 \\
			ESF & 0.432& 0.361& 0.285& 0.216 \\
			Fast Histogram & 0.612& 0.503& 0.384& 0.273 \\
			LocNet & \textbf{0.717}& \textbf{0.642}& \textbf{0.522}& \textbf{0.385} \\
			\hline
		\end{tabular}
	\end{center}
\end{table}

\begin{table}[t]
	\begin{center}
		\caption{ F$_{1}$max score on other sequences with $p=3m$}
		\begin{tabular}{cccccc}
			\hline
			Methods & Seq 02& Seq 05& Seq 06& Seq 08& Day 3-1 \\
			\hline
			Spin Image& 0.547& 0.550& 0.650& 0.566& \textbf{0.614} \\
			ESF & 0.461& 0.371& 0.439& 0.423& 0.373 \\
			Fast Histogram & 0.513& 0.569& 0.597& 0.521& 0.531\\
			LocNet & \textbf{0.702}& \textbf{0.689}& \textbf{0.714}& \textbf{0.664}& 0.607 \\
			\hline
		\end{tabular}
	\end{center}
\end{table}

The results of a place recognition algorithm are described using precision and recall metrics \cite{lowry2016visual}. We use the maximum value of F$_{1}$ score to evaluate different precision-recall relationships. With the ground truth provided by datasets, we consider that two places are the same if their Euclidean distance is below $p$. So different values of threshold $p$ determines the final  performances. We test different $p$ values on sequence 00 and the results are shown in TABLE \uppercase\expandafter{\romannumeral1}.

Besides, we set $p=3m$ and test the four methods on other sequences and  F$_{1}$ scores are shown in TABLE \uppercase\expandafter{\romannumeral2}. Obviously, the proposed LocNet achieves the best performance in most tests compared to other methods. 

\subsection{Localization Probability}

\begin{figure}[t]
	\centering
	\includegraphics[width=5.5cm]{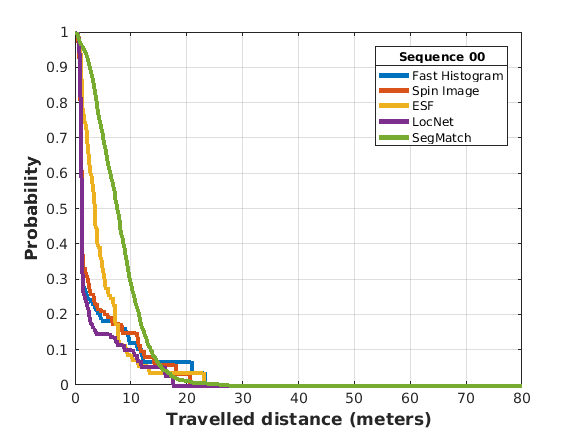}
	\caption{ Localization probability of traveling a given distance before localizing on the target global map of sequence 00. }
	\label{fig:location_prob}	
\end{figure}

\begin{figure}[t]
	\centering
	\subfloat[Sequence 00 (bird view)]{\includegraphics[width=3.3cm]{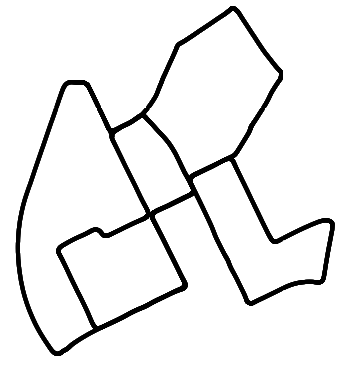}%
		\label{fig:origin}}
	\hfil
	\subfloat[Loop closure detection]{\includegraphics[width=3.3cm]{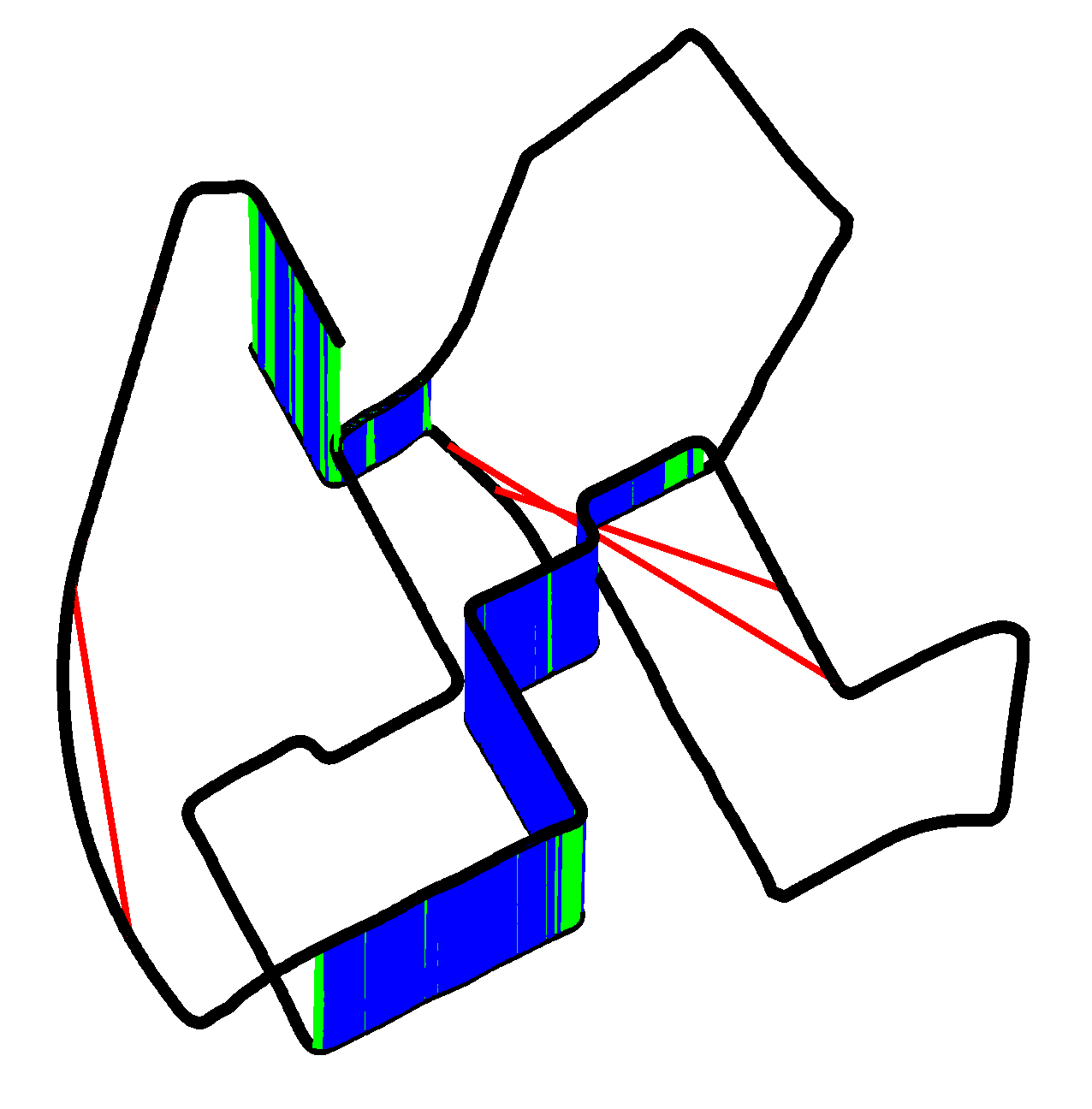}%
		\label{fig:loop}}
	\caption{ Place recognition detections on loop sections in sequence 00. In (b), red line stands for false positive match, green line for false negative and blue line for true positive. We add a growing height to each pose with respect to the time, so that the match can be visualized clearly.}
	\label{fig:drawLoop}
\end{figure}

In order to evaluate the performance of localizing, \cite{dube2017segmatch} shows the probability of traveling a given distance without successful localization in the target map. Thanks to the open source code of SegMatch\footnote{https://github.com/ethz-asl/segmatch}, we run 10 times in sequence 00 on the entire provided target map and present the average result. We record all the segmentation points as the successful localizations from the beginning to the end. As for the other four methods, we build the kd-tree based on the vectorized representations and look for the nearest pose when the current frame comes. If it is a true positive matching, we record it as a successful localization. There are no random factors in these four methods, so we run them only once. It is worth mentioning that SegMatch can return a full localization transformation using geometric verification, which is different from other place recognition algorithms.

As shown in Fig.~\ref{fig:location_prob}, the vehicle can localize itself within 20 meters 100$\%$ of the time using LocNet. It is nearly 98$\%$ for SegMatch and Spin Image, and 95$\%$ for the other two. The geometric methods and LocNets are based on one frame of LiDAR data, so they can achieve a higher percentage of the timing within 10 meters. While SegMatch relies on the semantic segmentation of the environments, so it needs an accumulation of 3D point clouds, thus resulting in a poorer performance in short traveling distance. Additionally, the loop closing results using LocNet on sequence 00 is shown as Fig.~\ref{fig:drawLoop}. These experiments all support that our system is effective to yield the correct topologically matching frame in the map.

\subsection{Global Localization Performance}

We demonstrate the performance of our global localization system in two ways: re-localization and position tracking. If a vehicle loses its location, re-localization ability is the the key to help it localize itself; and the coming position tracking achieve the localization continually.

\subsubsection{Re-localization}

\begin{figure}[t]
	\centering
	\subfloat[]{\includegraphics[width=4.3cm]{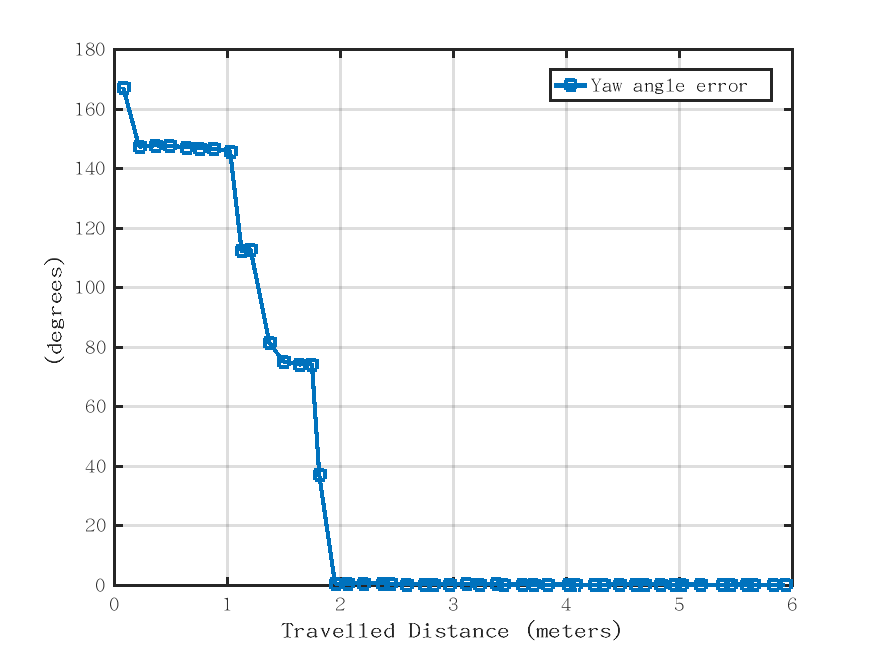}%
		\label{fig:degrees}}
	\hfil
	\subfloat[]{\includegraphics[width=4.3cm]{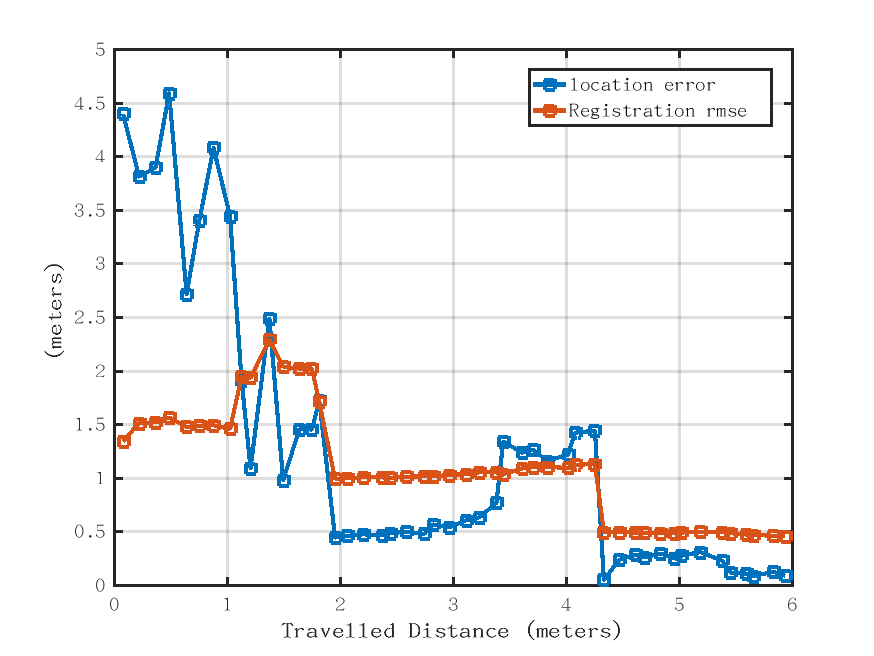}%
		\label{fig:meters}}
	\caption{(a) the convergence of yaw degree of re-localization (b) the decrease of location error and registration error. }
	\label{fig:convergency}
\end{figure}

The convergence rate of re-localization actually depends on the number of particles and the first observation they get in Monte-Carlo Localization. So it is hard to evaluate the convergence rate. We give a case study of re-localization using our vehicle in YQ21 dataset (see Fig.~\ref{fig:convergency}). The error of yaw angle and the positional angle are presented, together with the root mean square (rmse) error of registration, which is the Euclidean distance between the aligned point clouds $P$ and $P_{t}$.

Obviously the re-localization process can converges to a stable pose within 5m operation of the vehicle. And a change in location makes a change in the registration error. In multi-session datasets, the point clouds for localization are different from those frames forming the map, so the registration error can not be decreased to zero using ICP.

\subsubsection{Position Tracking}

\begin{figure}[!t]

	\centering
	\subfloat[Day3-1 location error]{\includegraphics[width=3.6cm]{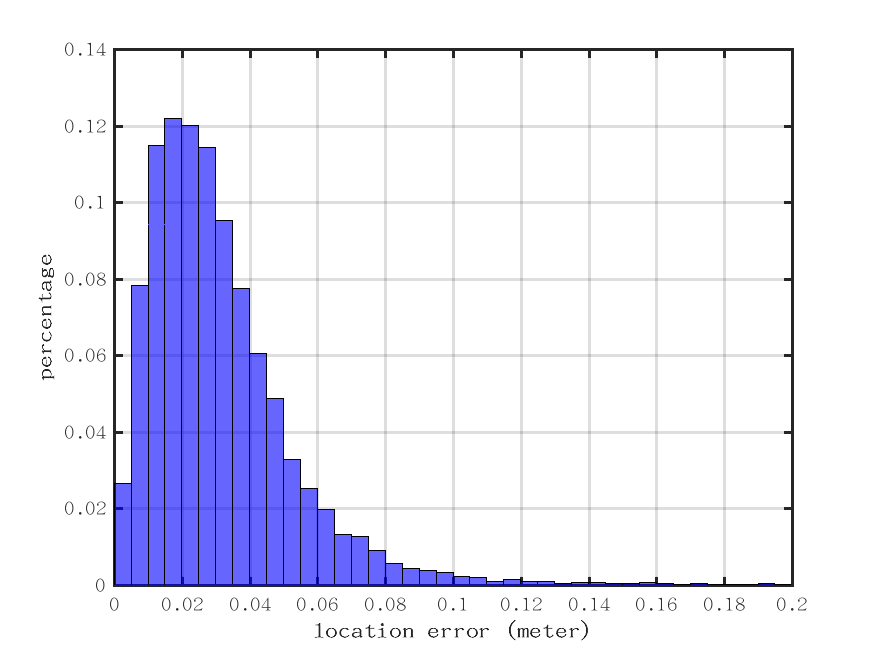}%
		\label{fig:090001_dis_h}}
	\hfil
	\subfloat[Day3-2 location error]{\includegraphics[width=3.6cm]{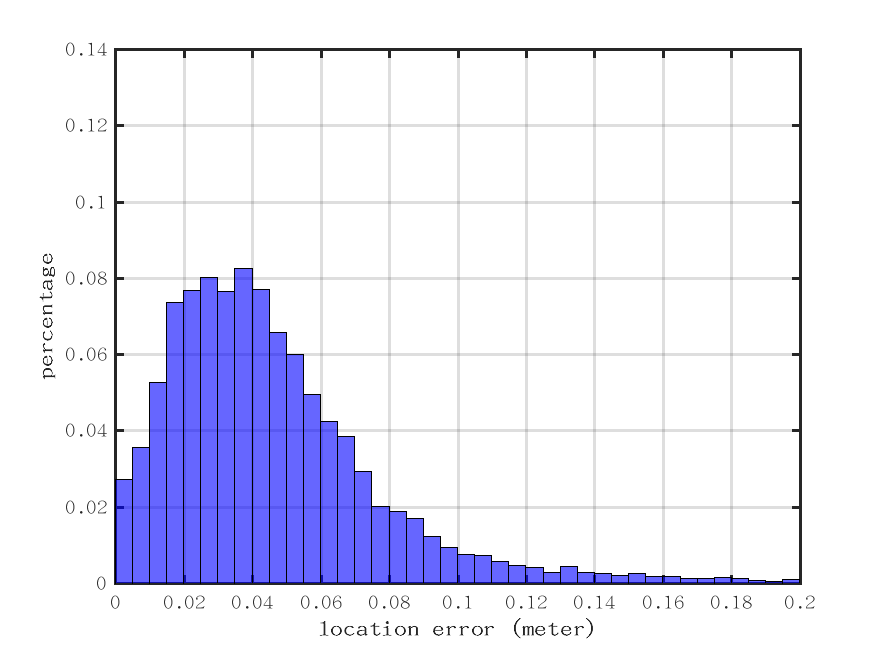}%
		\label{fig:090004_dis_h}} \\
	\subfloat[Day3-1 heading error]{\includegraphics[width=3.6cm]{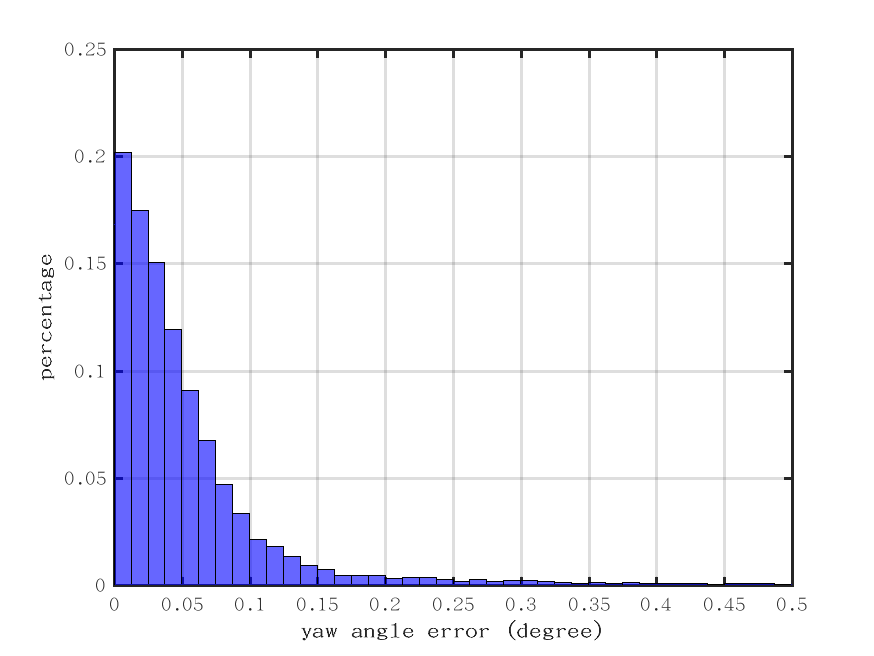}%
		\label{fig:090001_deg_h}}
	\hfil
	\subfloat[Day3-2 heading error]{\includegraphics[width=3.6cm]{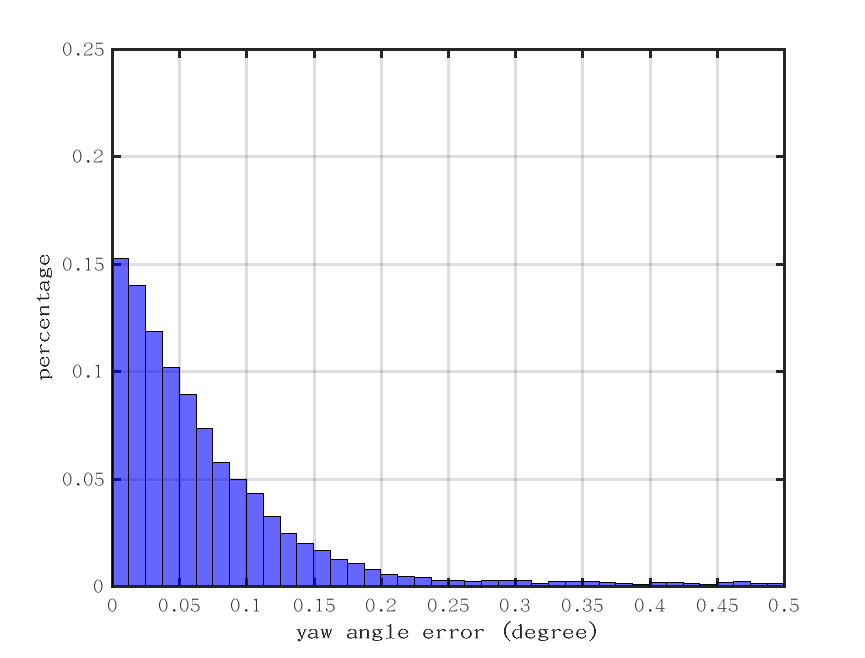}%
		\label{fig:090004_deg_h}}
	\caption{ Localization results of Day 3-1 and Day 3-2 on the global map of Day 1. (a) and (b): the distributions of location errors. (c) and (d): the distributions of heading errors (yaw angle errors).}
	\label{fig:position_tracking}
\end{figure}

Once the vehicle is re-localized, the global localization system can achieve the position tracking of the vehicle. The range-only observations provide reliable information of vehicle global pose, and ICP algorithm can help the vehicle localize more accurately. We test both Day 3-1 and Day 3-2 and the results are presented in  Fig.~\ref{fig:position_tracking}.

Based on the statistics analysis, our global localization system can achieve high accuracy of position tracking. 93.0$\%$ of location errors are below 0.1m in Day 3-1, and 88.9$\%$ in Day 3-2; for rotational errors, 93.1$\%$ of heading errors are below 0.2$^{\circ}$ in Day 3-1, and 90.7$\%$ in the afternoon of Day 3-2. The negligible errors cause no effect for autonomous navigation actions of the vehicle.
%%%%%%%%%%%%%%%%%%%%%%%%%%%%%%%%%%%%%%%%%%%%%%%%%%%%%%%%%%%%%%%%%%%%%%%%%%%%%%%%

\section{Conclusion}

This paper presents a global localization system in 3D point clouds for mobile robots or vehicles, evaluated on different datasets. The global localization method is based on the learned representations by LocNet in Euclidean space, which are used to build the necessary global prior map. In the future, we would like to achieve the global localization in the dynamic environments for mobile vehicles.

%%%%%%%%%%%%%%%%%%%%%%%%%%%%%%%%%%%%%%%%%%%%%%%%%%%%%%%%%%%%%%%%%%%%%%%%%%%%%%%%

%\section*{ACKNOWLEDGMENT}

%%%%%%%%%%%%%%%%%%%%%%%%%%%%%%%%%%%%%%%%%%%%%%%%%%%%%%%%%%%%%%%%%%%%%%%%%%%%%%%%
\bibliographystyle{IEEEtran}
\bibliography{root}

\end{document}